\documentclass[sigconf]{acmart}
\AtBeginDocument{%
  }

\setcopyright{acmlicensed}
\copyrightyear{2018}
\acmYear{2018}
\acmDOI{XXXXXXX.XXXXXXX}

\acmConference[Conference acronym 'XX]{Make sure to enter the correct
  conference title from your rights confirmation emai}{June 03--05,
  2018}{Woodstock, NY}
\acmISBN{978-1-4503-XXXX-X/18/06}

\begin{document}

%%
%% The "title" command has an optional parameter,
%% allowing the author to define a "short title" to be used in page headers.
\title{ EGEAN: An Exposure-Guided Embedding Alignment Network for Post-Click Conversion Estimation}

%%
%% The "author" command and its associated commands are used to define
%% the authors and their affiliations.
%% Of note is the shared affiliation of the first two authors, and the
%% "authornote" and "authornotemark" commands
%% used to denote shared contribution to the research.
\author{Guoxiao Zhang}
\authornote{Both authors contributed equally to this research.}
\email{zhangguoxiao@meituan.com}
\orcid{1234-5678-9012}

\author{Huajian Feng}
\authornotemark[1]
\email{299145@whut.edu.cn}
\affiliation{%
  \institution{Meituan}
  \city{Beijing}
  \country{China}
}
\author{Yadong Zhang}
\affiliation{%
  \institution{Meituan}
  \city{Beijing}
  \country{China}}
\email{zhangyadong05@meituan.com}

\author{Yi Wei}
\affiliation{%
  \institution{Meituan}
  \city{Beijing}
  \country{China}}
\email{weiyi20@meituan.com}

\author{Qiang Liu}
\affiliation{%
  \institution{Meituan}
  \city{Beijing}
  \country{China}}
  \email{liuqiang43@meituan.com}

%%
%% By default, the full list of authors will be used in the page
%% headers. Often, this list is too long, and will overlap
%% other information printed in the page headers. This command allows
%% the author to define a more concise list
%% of authors' names for this purpose.
\renewcommand{\shortauthors}{Trovato et al.}

%%
%% The abstract is a short summary of the work to be presented in the
%% article.
\begin{abstract}
Accurate post-click conversion rate (CVR) estimation is crucial for online advertising systems. Despite significant advances in causal approaches designed to address the Sample Selection Bias problem, CVR estimation still faces challenges due to Covariate Shift. Given the intrinsic connection between the distribution of covariates in the click and non-click spaces, this study proposes an Exposure-Guided Embedding Alignment Network (EGEAN) to address estimation bias caused by covariate shift. Additionally, we propose a Parameter Varying Doubly Robust Estimator with steady-state control to handle small propensities better. Online A/B tests conducted on the Meituan advertising system demonstrate that our method significantly outperforms baseline models with respect to CVR and GMV, validating its effectiveness. Code is available: \url{https://github.com/hydrogen-maker/EGEAN}.
\end{abstract}

%%
%% The code below is generated by the tool at http://dl.acm.org/ccs.cfm.
%% Please copy and paste the code instead of the example below.
%%


%%
%% Keywords. The author(s) should pick words that accurately describe
%% the work being presented. Separate the keywords with commas.
%% A "teaser" image appears between the author and affiliation
%% information and the body of the document, and typically spans the
%% page.

%%
%% This command processes the author and affiliation and title
%% information and builds the first part of the formatted document.
\maketitle
\section{Introduction}
\begin{figure}[ht]
    \centering
    \includegraphics[width=\linewidth]{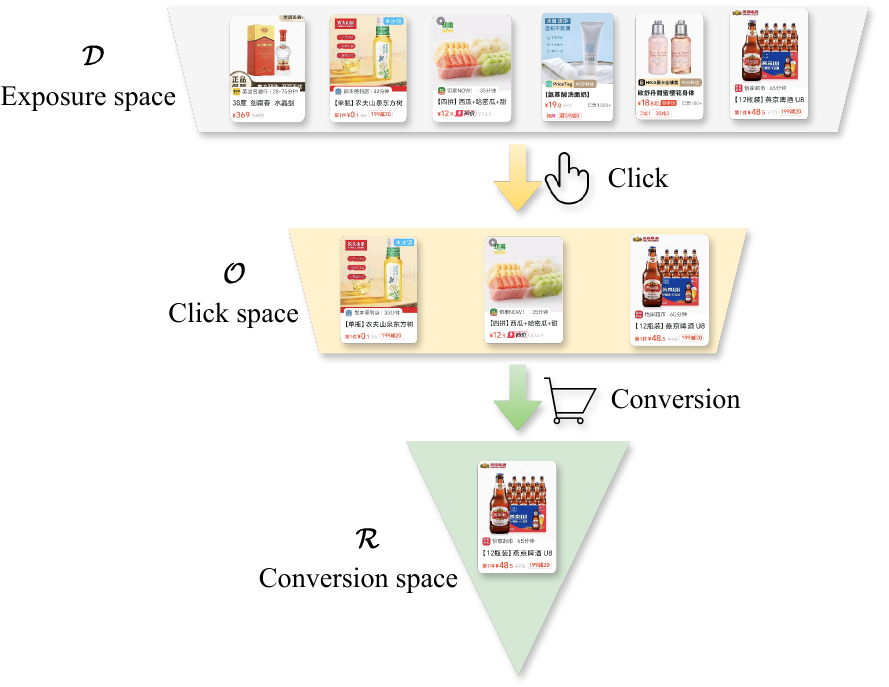}
    \caption{Illustration of user behaviors in Meituan online advertising.}
    \label{fig:Figure1}
\end{figure}

The post-click conversion rate (CVR) is essential for revenue growth in online advertising systems\cite{zhang2024adversarial}. The CVR task faces significant challenges because it is trained in the click space while predictions are made in the impression space, which inevitably encounters Sample Selection Bias(SSB) problem: the missing conversion labels follow a missing-not-at-random(MNAR) pattern, leading to a misalignment between the event distributions in the training space and the inference space. 
\autoref{fig:Figure1} illustrates this bias and its formation process.

Currently, some causal approaches are proposed to adapt for the biased estimates from NMAR data\cite{zhang2020large,wang2022escm2,zhu2023dcmt,zhang2024adversarial,huang2024utilizing,su2024ddpo,liu2024ranking}. The proof of the unbiasedness of these IPS and DR estimators is based on the assumption that the distribution of covariates is consistent, which has been pointed out to be false in AECM\cite{zhang2024adversarial}. AECM\cite{zhang2024adversarial} first introduces an adversarial method of Domain Adaptation(DA) to address the covariate shift problem between click and unclick spaces. However, the adversarial method does not always work well on small datasets and is relatively challenging to optimize\cite{zhao2020review}. Besides, the general method of DA ignores an intrinsic connection between the distribution of covariates in the click and non-click space.

Inspired by the fact that the exposure event is the shared part of click and non-click events, indicating an intrinsic connection between the distribution of covariates in the click and non-click spaces, we discard the conventional adversarial training solutions and propose an Exposure-Guided Embedding Alignment Network (EGEAN). Specifically, we first align the distribution of covariates to the exposure space through an exposure task. Then, we add a task-personalized network to enhance the efficiency of CTR  and CVR estimation with a loss function that minimizes the distribution gap. Besides, to handle small propensities better, we draw on the methods from StableDR\cite{li2023stabledrstabilizeddoublyrobust} and propose a Parameter Varying DR(PVDR) Estimator with steady-state control. 
Our contributions include:
\begin{itemize}
\item Proposing an exposure-guided embedding alignment network to address estimation bias caused by covariate shift.
\item Introducing a parameter varying doubly robust estimator with steady-state control to handle small propensities better.
\item Successfully deploying this method on the Meituan online advertising system, achieving 5.94\% rise in CVR during online A/B tests.
\end{itemize}

\section{PROPOSED METHOD}
\begin{figure}[ht]
    \centering
    \includegraphics[width=\linewidth]{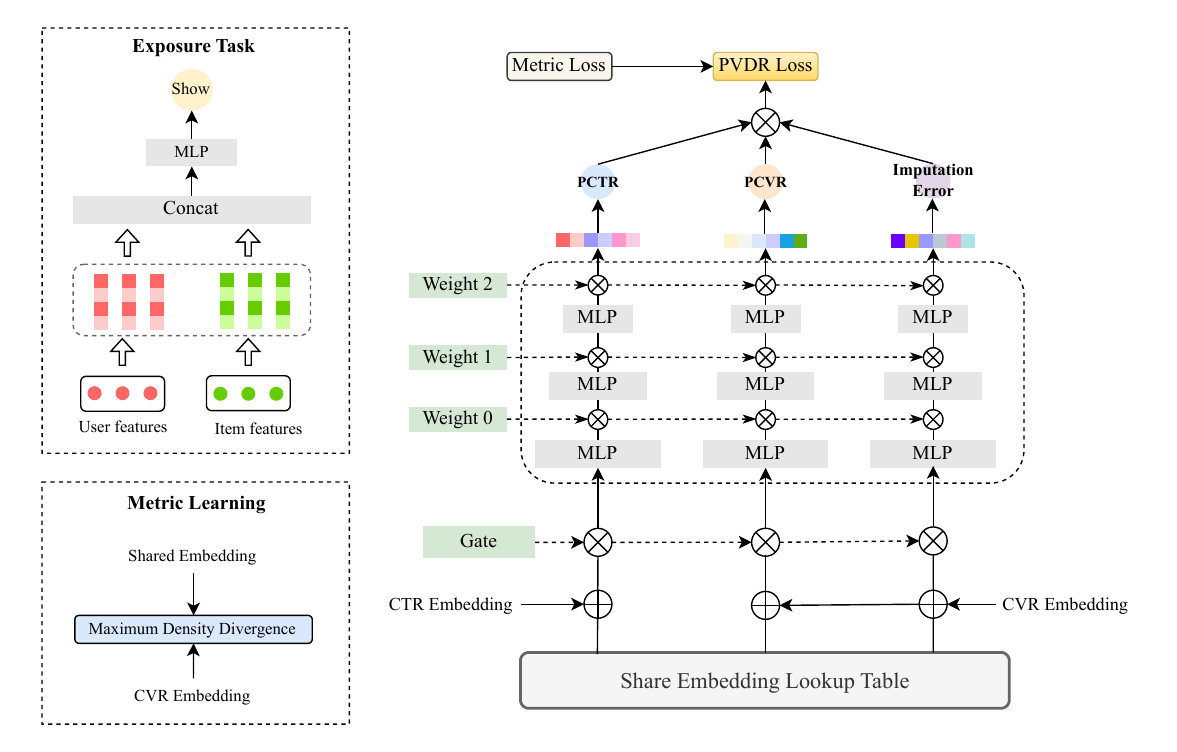}
    \caption{The overall framework of EGEAN.}
    \label{fig:framework}
\end{figure}

% 重写这部分，做了什么，解决了什么问题
In this section, we describe our proposed \textbf{EGEAN} as shown in Figure \ref{fig:framework} and PVDR estimator.

\subsection{Exposure-
Guided Embedding Alignment Network}
EGEAN comprises three fundamental mechanisms: 1) an exposure probability prediction task. 2) a task personalized network. 3) a metric learning module.
\subsubsection{Exposure Probability Prediction Task.}
% 改名字：generator
We propose an exposure probability prediction task to capture the common features of both clicked and unclicked events, as shown in Figure \ref{fig:framework}. For this task, positive samples are exposed samples, while negative samples are generated using in-batch negative sampling. The network concatenates the original embeddings of users and
items, which are then processed through a Multi-Layer Perceptron
(MLP):
\begin{equation}
    \hat{y}^{i}_{exp} = \sigma(\text{MLP}(\mathbf{x}_{u,i}))\end{equation}
where \( y^{i}_{\text{exp}} \in (0, 1) \) is the predicted exposure probability of the \( i \)-th original item, \( \sigma \) is the sigmoid function, and $\mathbf{x}_{u,i}$ indicates the concatenation of user features and item features. We use cross-entropy loss to optimize this network with the exposure label.

\subsubsection{Task Personalized Network.}
After the pretraining phase, our objective is to enhance the efficiency of multi-task learning by incorporating a task-personalized network. This network is structured around three integral components:1) At the input layer, we incorporate LoRA to fine-tune the embeddings.2)At the representation layer, we add an embedding personalized network.3)At the MLP layer, we introduce a parameter personalized network.

\textbf{LoRA}.First, drawing inspiration from the Low-Rank Adaptation (LoRA)\cite{hu2021lora}. We finetune the pre-trained embeddings for CTR task and CVR task using Low-Rank adaptors. Specifically, for a pre-trained embeddings $W \in \mathbb{R}^{d\times k}$, we constrain its update by representing the latter with a low-rank decomposition $W + \Delta W = B A$, where $B \in \mathbb{R}^{d\times r}$, $A \in \mathbb{R}^{r\times k}$, and the rank $r \ll \min(d, k)$. During training, $W$ is frozen and does not receive gradients, while the $A$ and $B$ contain trainable parameters. 

\textbf{PNet}.Following \cite{chang2023pepnet},  We utilize task-side features \(E(\mathcal{F}_{d}) \in \mathbb{R}^{k}\) as the input of EPNet, including task ID. \(\mathbf{U}_{ep}\) is the Gate NU of EPNet in the embedding layer, and its output \(\delta_{task} \in \mathbb{R}^{e}\) is given by:
\begin{equation}
\delta_{task} = \mathbf{U}_{ep}(E(\mathcal{F}_{d}) \oplus (\varnothing(\mathbf{E}))),
\end{equation}

where we concatenate the general embedding \(\mathbf{E} \in \mathbb{R}^{e}\) with the input, but without using gradient backpropagation, denoted as \(\varnothing(*)\). Next, we employ the external Gate NU to perform the personalized transformation on embedding \(\mathbf{E}\) without altering the original embedding layer. The transformed embedding is:
\begin{equation}
\mathbf{O}_{ep} = \boldsymbol{\delta}_{task} \otimes \mathbf{E},
\end{equation}
where \(\mathbf{O}_{ep} \in \mathbb{R}^{e}\), and \(\otimes\) denotes the element-wise product.

\textbf{PPNet}
Following [1], we propose PPNet to modify DNN parameters in multi-task learning. We use task-side features (\(\mathcal{F}_{t}\)) as priors for PPNet, such as task ID, etc. Specifically, the detailed structure of PPNet is as follows:

% \begin{equation}
% \mathbf{O}_{prior} &= E(\mathcal{F}_{t}), \\
% \boldsymbol{\delta}_{task} &= \mathbf{U}_{pp}(\mathbf{O}_{prior} \oplus (\varnothing(\mathbf{O}_{ep}))).
% \end{equation}

We concatenate the output of EPNet \(\mathbf{O}_{ep}\) with the personalized prior \(\mathbf{O}_{prior}\) as the input of \(\mathbf{U}_{pp}\), which is the Gate NU in PPNet. To avoid affecting the embedding updated in EPNet, we perform the operation of stop gradient on \(\mathbf{O}_{ep}\).

Furthermore, we integrate PPNet into all DNN layers to fully personalize DNN parameters, balancing targets with different sparsity for different users in multiple tasks, formulated as follows:

\begin{equation}
\mathbf{O}^{(1)}_{pp} = \boldsymbol{\delta}_{task} \otimes \mathbf{H}^{(l)},
\end{equation}

\begin{equation}
\mathbf{H}^{(l+1)} = f(\mathbf{O}^{(1)}_{pp} \mathbf{W}^{(l)} + \mathbf{b}^{(l)}), \quad l \in \{1, \dots, L\},
\end{equation}

where \(L\) is the number of DNN layers of task towers and \(f\) is the activation function.
\subsubsection{Metric learning.}

% 思想是在曝光emb个性化时控制
Finetuning for CVR task with the embedding personalized network may lead to discrepancies in the distribution of clicks and non-click spaces. To address this issue, we employ metric learning to compute the distributional distance. Specifically, we calculate the distance between the CVR and shared embedding, which we designate as the metric loss. This merit loss is then incorporated into the final model loss. The model can better transfer knowledge learned from the non-click space to the click space by minimizing the metric loss. Following \cite{geng2011daml}, we utilize Maximum Mean Discrepancy (MMD) to measure the distributional distance. The formula is typically expressed as:

\begin{equation}
\text{MMD}^2(\mathcal{F}, p, q) = \left\| \mathbb{E}_{p}[\phi(x)] - \mathbb{E}_{q}[\phi(y)] \right\|^2_{\mathcal{H}}
\end{equation}

where $\mathcal{F}$ is a class of functions, $p$ and $q$ are two distributions, $\phi$ is a feature mapping to a Reproducing Kernel Hilbert Space (RKHS), $\mathbb{E}_{p}[\phi(x)]$ denotes the expectation of the feature mapping of samples drawn from distribution $p$, and $\mathbb{E}_{q}[\phi(y)]$ denotes the expectation of the feature mapping of samples drawn from distribution $q$.

\subsection{Parameter Varying DR Estimator}

\subsubsection{Problem Definition}
In CVR prediction, the model takes user-item features $x_{u,i}$ and outputs a conversion probability $\hat{r}_{u,i}$. We denote the prediction matrix by $\hat{\mathbf{R}} \in \mathbb{R}^{n \times m}$, where $\hat{r}_{u,i} \in [0, 1]$. If $\mathbf{R}$ were fully observed in both click space $\mathcal{O}$ and non-click space $\mathcal{N}$, the ideal loss function would be:
\begin{equation}
\mathcal{L}_{\text{ideal}} = \mathbb{E}(\mathbf{R}, \hat{\mathbf{R}}) = \frac{1}{|\mathcal{D}|} \sum_{(u,i) \in \mathcal{D}} \delta(r_{u,i}, \hat{r}_{u,i}),
\end{equation}
where $\delta(r_{u,i}, \hat{r}_{u,i}) = -r_{u,i} \log(\hat{r}_{u,i}) - (1 - r_{u,i}) \log(1 - \hat{r}_{u,i})$. However, $\mathcal{L}_{\text{ideal}}$ is theoretical, as non-click space labels are unobserved. The bias between a model $M$'s loss $\mathcal{L}_{\mathcal{M}}$ and $\mathcal{L}_{\text{ideal}}$ is:
\begin{equation}
\mathrm{Bias}[\mathcal{L}_{\mathcal{M}}] = |\mathcal{L}_{\mathcal{M}} - \mathcal{L}_{\text{ideal}}|.
\end{equation}
Conversions are only observed in $\mathcal{O}$, leading to the naive estimator's loss:
\begin{equation}
\mathcal{L}_{\text{naive}} = \frac{1}{|\mathcal{O}|} \sum_{(u,i) \in \mathcal{O}} o_{u,i} \delta(r_{u,i}, \hat{r}_{u,i}).
\end{equation}

\subsubsection{PVDR Estimator.}
We present our PVDR Estimator, which incorporates techniques from StableDR[7]. This approach consists of three components: training an imputation model, implementing a steady-state control condition, and developing a PVDR Estimator.

\textbf{Imputation model.}
Initially,we train an imputation model $\hat{e}_{ui} = \phi_{\theta}(\mathbf{x}_{u,i})$.The imputation model, parameterized by $\phi$, aims to estimate the CVR prediction error $\hat{e}_{u,i}$ with $\mathbf{x}_{u,i}$.The loss of imputation model is  given as
\begin{equation}
\hat{\mathcal{L}} = \frac{1}{|\mathcal{D}|} \sum_{(u,i) \in \mathcal{D}} \hat{e}_{u,i}
\end{equation}

\textbf{Steady-state control condition.}
% 此处改为稳态条件，内容修改为：此处提出了一个稳态控制条件，具体而言：
We propose an ingenious formula to determine that PVDR is unbiased, which we refer to as the steady-state control condition. Specifically,
\begin{equation}
\lambda +(1-\lambda)A=B
\end{equation}
where $\lambda$ is a hyperparameter, $A = \frac{1}{|\mathcal{D}|} \sum\limits_{(u,i) \in \mathcal{D}} \frac{o_{u,i}}{\hat{p}_{u,i}}$,\\
$B =\sum_{(u,i)\in \mathcal{D}} \frac{o_{u,i} \hat{e}_{u,i}}{\hat{p}{u,i}}/\sum_{(u,i)\in \mathcal{D}} \hat{e}_{u,i}$

\textbf{PVDR Estimator.}
The PVDR estimator is given as
\begin{equation}
\mathcal{L}_{PVDR} = \frac{\sum\limits_{(u,i) \in \mathcal{D}} \frac{o_{u,i} e_{u,i}}{\hat{p}_{u,i}}}{\lambda \left |\mathcal{D}  \right | +(1-\lambda ) \sum\limits_{(u,i) \in \mathcal{D}} \frac{o_{u,i}}{\hat{p}_{u,i}}}
\end{equation}

When the steady-state condition is met, PVDR is a doubly robust estimator, and the CVR estimation is unbiased. 
\begin{equation}
\begin{aligned}
\lambda +(1-\lambda)\frac{1}{|\mathcal{D}|} \sum\limits_{(u,i) \in \mathcal{D}} \frac{o_{u,i}}{\hat{p}_{u,i}}=\sum_{(u,i)\in \mathcal{D}} \frac{o_{u,i} \hat{e}_{u,i}}{\hat{p}{u,i}}/\sum_{(u,i)\in \mathcal{D}} \hat{e}_{u,i}
\end{aligned}
\end{equation}
Obviously, StableDR\cite{li2023stabledr}is a special case of PVDR when $\lambda=1$
, and it degenerates to IPS when $\lambda=0$.
% \subsubsection{Bias and Variance Analysis of ADR Estimator}
% 1.加无偏性证明
% 2.是一个DR估计量
% 3.方差更优（SDR是一个special case，倾向得分较低时SDR方差优于IPS和传统DR）
Due to the introduction of parameter 
$\lambda$, it is evident that the variance of PVDR outperforms SDR when the propensity score is relatively small.

\section{Experiments}
We conduct extensive experiments on the public datasets and one private industrial dataset. The statistics of the processed datasets are reported in Table \ref{tab:dataset_stats}.
\begin{itemize}
    \item \textbf{Ali-CCP} :The public dataset Ali-CCP \cite{ma2018entire} gathered from real-world traffic logs of the recommender system in Taobao.The Ali-CCP contains 400 thousand users and 4.3 million items, as well as over 80 million user-item interactions.
    \item \textbf{Meituan} : Three months of search logs in recommend ad system in Meituan are to generate the examples and corresponding features, ranging from July. 2024 to Oct. 2024
\end{itemize}
\begin{table}[h]
\centering
\caption{Statistics of datasets. B denotes billions, while M and K
signify millions and thousands, respectively. Conv is short for conversion.}
\label{tab:dataset_stats}
\begin{tabular}{lrrrrr}
\toprule
\textbf{Dataset} & \textbf{\#Users} & \textbf{\#Items} & \textbf{\#Exposure} & \textbf{\#Click}& \textbf{\#Conv}\\
\midrule
Ali-CCP & 0.4M & 4.3M & 85.3M& 3.3M & 18.4K  \\
Meituan & 97.5M & 0.1B & 2.1B & 33.5M &4.8M\\
\bottomrule
\end{tabular}
\end{table}
\subsubsection{Evaluation Metric}
We apply widely used AUC (Area Under ROC) as the primary metric to evaluate the performance.
\subsubsection{Baseline Methods}
We compare our proposed framework with the following baselines.
\begin{itemize}
    \item \textbf{ESMM}\cite{ma2018entire} : It learns CVR utilizing a CTR task and a CTCVR task which is a non-causal estimator.
    \item \textbf{MTL-DR}\cite{10.1145/3366423.3380037} : It deploys the DR estimator within a multi-task framework to debias more robustly.
    \item \textbf{DDPO} \cite{su2024ddpo}: It incorporates the IPW estimator to regularize ESMM’s CVR estimation.
    \item \textbf{DCMT} \cite{zhu2023dcmt}: It optimizes two opposite CVR towers in the click space and the non-click space.
    \item \textbf{AECM}\cite{zhang2024adversarial}: It optimizes two opposite CVR towers in the click space and the non-click space.
\end{itemize}
\subsubsection{Implementation Details}
For a fair comparison, we adopt the following settings for all methods: The batch size is set to 1024 and we employ the Adam\cite{kingma2017adammethodstochasticoptimization} optimizer with a learning rate of $lr = 1 \times 10^{-3}$ and weight decay of $\beta = 1 \times 10^{-3}$. The activation functions involved in two discriminators make use of LeakyReLU\cite{maas2013rectifier} with a negative slope set to 0.2. The embedding size is set to 5 for the Ali-CCP dataset and 8 for the Meituan dataset.Xavier initialization\cite{pmlr-v9-glorot10a} is used here to initialize the parameters.
% table2更换为我们的消融实验结果 ，策略+曝光+模块
\subsection{Overall Performance}
In this section, we compare the performance of our proposed EGEAN with the baselines on the above two datasets. We consider two distinct tasks, i.e., CVR and CTCVR, and record data in the entire impression space $\mathcal{D}$ with the metric AUC for evaluation. The overall results are shown in Table \ref{tab:result1}.
\begin{table}[ht]
\centering
\caption{The performance comparison of different models. All results are reported in terms of AUC. EGEAN+DR is a combination approach formed by applying our method to the base estimator DR. The best and the second results are marked with Bold and Underline.}
\label{tab:result1}
\begin{tabular}{cc|cc|cc}
\toprule
  \multirow{2}*{\textbf{Model}}&  & \multicolumn{2}{c|}{\textbf{Ali-CCP}} & \multicolumn{2}{c}{\textbf{Meituan}} \\
 &  & \textbf{CVR}& \textbf{CTCVR} & \textbf{CVR}& \textbf{CTCVR} \\ \midrule
ESMM  &  & 0.5813 & 0.5799 & 0.6507 & 0.6582 \\
MTL-DR &  & 0.6127 & 0.6117 & 0.6713 & 0.6789 \\
DDPO  &  & \underline{0.6401} & 0.6336 & \underline{0.6802} & 0.6797 \\
DCMT  &  & 0.6398 & \underline{0.6384} & 0.6679 & 0.6702 \\
AECM &  & 0.6192 & 0.6133 & 0.6791 & \underline{0.6825} \\ \midrule
EGEAN+DR &  & 0.6527 &0.6455& 0.6940 & 0.6971 \\ 
EGEAN+PVDR(Ours)&  & \textbf{0.6604} & \textbf{0.6550} & \textbf{0.7093} & \textbf{0.7132} \\ \bottomrule
\end{tabular}
\end{table}

As illustrated in Table 2, EGEAN demonstrates a markedly superior performance compared to all baselines for the CVR prediction task across both datasets. Compared with the best baseline DDPO, EGEAN+DR achieves \textbf{3.17\%} and \textbf{4.28\%} AUC improvements over the Ali-CCP and Meituan datasets, respectively. EGEAN combines the strength of PVDR and embedding alignment network to eliminate estimation bias. The significant improvement indicates the positive effects of our method
\subsection{Ablation Study}
To analyze the contributions of each component in our model, we conduct the ablation studies presented in Table \ref{tab:ablation study}.
The framework consists of three integral components: an exposure network, a task personalization network, and a metric learning module.
As shown in Figure \ref{fig:framework}, three degraded versions suffer a decrease over the two datasets compared to the original EGEAN. This indicates the effectiveness of our well-designed user task personalization modeling and distribution alignment strategy.

\begin{table}[ht]
\centering
\caption{Results of the ablation study. w/o is short for without. EN is short for the exposure network. TPN and ML are short for the task personalization network and metric learning task, respectively.}
\label{tab:ablation study}
\begin{tabular}{cc|cc|cc}
\toprule
 \multirow{2}*{\textbf{Methods}}&  & \multicolumn{2}{c|}{\textbf{Ali-CCP}} & \multicolumn{2}{c}{\textbf{Meituan}} \\
 &  & \textbf{CVR}& \textbf{CTCVR} & \textbf{CVR}& \textbf{CTCVR} \\ \midrule
EGEAN w/o EN &  & 0.6274 & 0.6323 & 0.6479 & 0.6582 \\
EGEAN w/o TPN &  & 0.5958 & 0.6065 & 0.6218 & 0.6488 \\
EGEAN w/o ML &  & 0.6364 & 0.6316 & 0.6869 & 0.6887 \\ \midrule
EGEAN(full model) &  & \textbf{0.6604} & \textbf{0.6550} & \textbf{0.7093} & \textbf{0.7132} \\  \bottomrule
\end{tabular}
\end{table}

\subsection{Case Study}
\begin{figure}[ht]

    \begin{minipage}{0.49\linewidth}%可修改0.49为其他比例，调整大小
        \vspace{3pt}
        \centerline{\includegraphics[width=\textwidth]{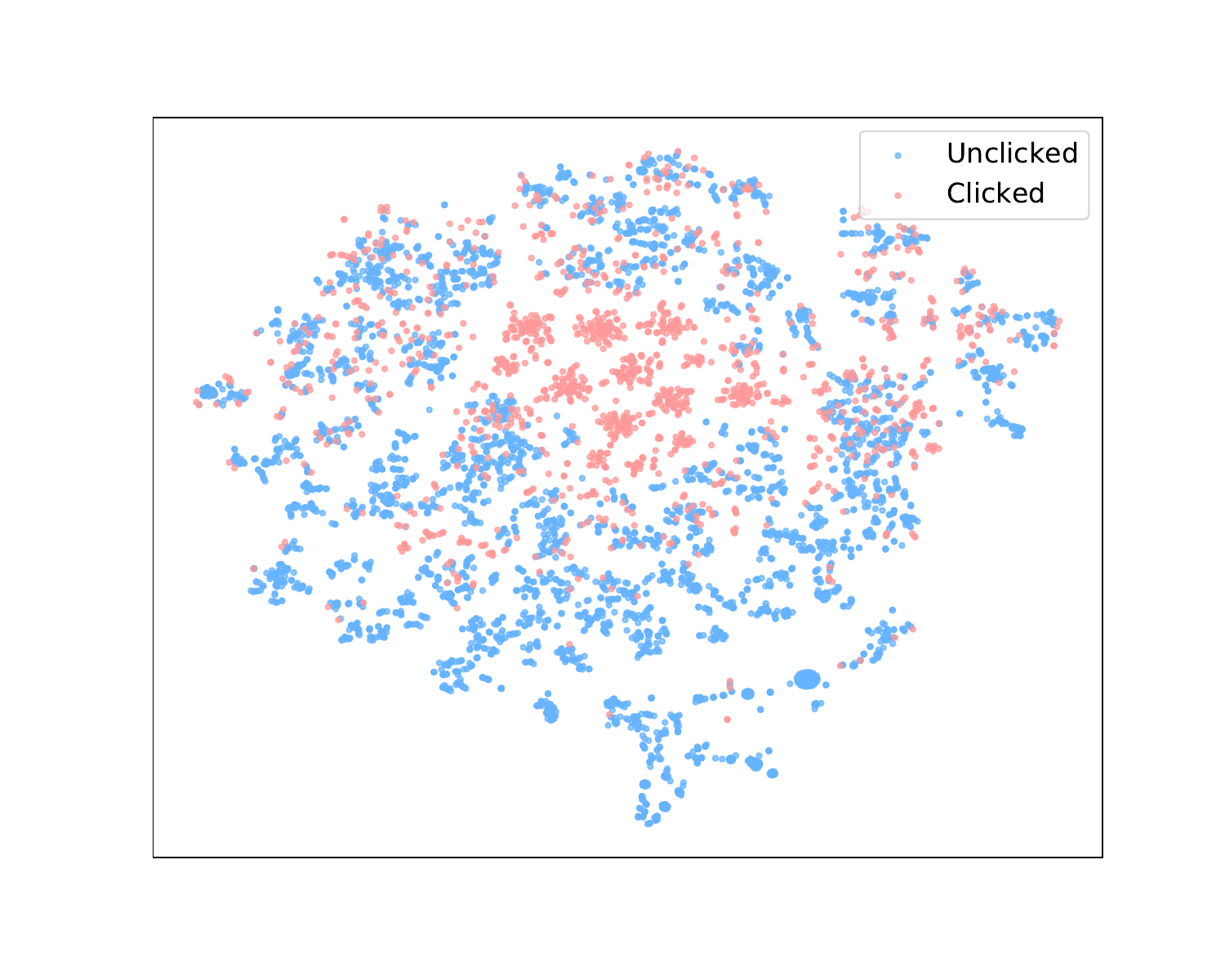}}
    \end{minipage}
    \begin{minipage}{0.49\linewidth}
        \vspace{3pt}
        \centerline{\includegraphics[width=\textwidth]{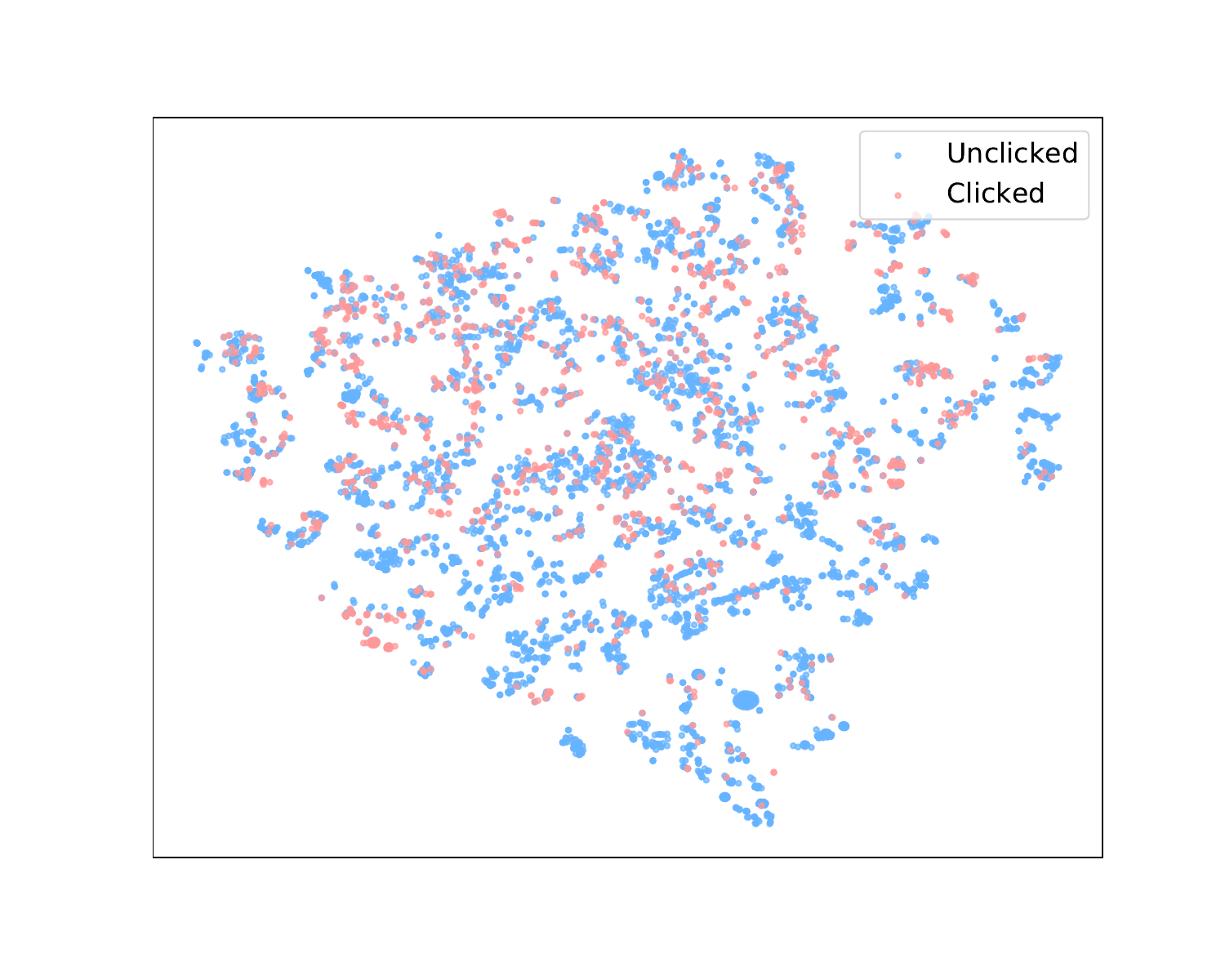}}
    \end{minipage}

    \caption{The t-SNE visualization of feature distributions for TSN modules on the Industrial dataset Meituan. Blue and cherry blossom representing the unclicked and clicked events, respectively.}
    \label{distribution}
\end{figure}

Fig \ref{distribution}(a) and Fig \ref{distribution}(b) display the characteristic distribution of embeddings for the Meituan dataset. The visualizations in these figures vividly demonstrate that in the absence of the TSN module, there is a notable separation between the distributions of clicked events (represented by cherry blossom dots) and unclicked events (blue dots). Conversely, upon integrating the TSN module, the clicked and unclicked events are fully integrated into the same feature space, leading to their complete intermingling. These observations provide a degree of confirmation for the effectiveness of our proposed framework and the validity of our experimental results.

%  \begin{figure}[htb]
% 	\begin{center}
% 	\subfigure[]{
%  		\label{}
%  		\vspace{-1mm}
%  		\includegraphics[width=0.21\textwidth]{tsne_visualization_before.pdf}}
%  		%\hfill
%  	\subfigure[]{
%  		\label{}
%  		\vspace{-1mm}
%  		\includegraphics[width=0.21\textwidth]{tsne_visualization_after.pdf}}
% 	\caption{The t-SNE visualization of feature distributions for TSN modules on the Industrial dataset Meituan. Blue and cherry blossom representing the unclicked and clicked events, respectively.}
% 	\label{distribution}
% 	\end{center}
% 	\vspace{-2mm}
% \end{figure}

\subsection{Online A/B Test}
From 2024-10-01 to 2024-10-07, we conducted online A/B testing on the online advertising system in Meituan to validate the proposed EGEAN. As shown in Table 3, compared to the DDPO model (our last product model), EGEAN achieves 5.94\% CVR and 6.29\% GMV improvement. Now, EGEAN serves the main traffic in the advertising system in Meituan.

\section{Conclusions}

In this paper, we introduce a novel model, EGEAN, that tackles the covariate shift challenges prevalent in CVR estimation. Through the seamless integration of a parameter varying doubly robust estimator ,EGEAN has exhibited remarkable enhancements compared with existing baseline models across diverse datasets . Our model notably increases system GMV by 6.29\% and CVR by 5.94\% on Meituan online advertising systems.This approach opens new research possibilities in CVR estimation, and we hope our work will inspire further exploration in this area.
%%
%% The next two lines define the bibliography style to be used, and
%% the bibliography file.
\bibliographystyle{ACM-Reference-Format}
\bibliography{sample-base}

%%
%% If your work has an appendix, this is the place to put it.
\appendix
\end{document}